\documentclass[a4paper]{article}

\usepackage{INTERSPEECH2020}
\usepackage{hyperref}
\usepackage[ruled,vlined]{algorithm2e}
\usepackage{amsmath}
\usepackage{subfig}

\usepackage{tikz}
\usepackage{float}
\usepackage{enumitem}
\usetikzlibrary{bayesnet}
\usetikzlibrary{automata, positioning}
\usetikzlibrary{arrows}

\title{Modeling Global Body Configurations in American Sign Language}
 \name{Nicholas Wilkins$^1$, Beck Cordes Galbraith$^2$, Ifeoma Nwogu$^1$}
 \address{
   $^1$Rochester Institute of Technology, Rochester, NY, USA
   $^2$Sign-Speak, USA
  }
\email{npw3202@rit.edu, beck.cordes.galbraith@gmail.com, ion@cs.rit.edu}

\begin{document}

\maketitle
\begin{abstract}

In this paper we consider the problem of computationally representing American Sign Language (ASL) phonetics. We specifically present a computational model inspired by the sequential phonological ASL representation, known as the Movement-Hold (MH) Model. Our computational model is capable of not only capturing ASL phonetics, but also has generative abilities. We present a Probabilistic Graphical Model (PGM) which explicitly models holds and implicitly models movement in the MH model. For evaluation, we introduce a novel data corpus, ASLing, and compare our PGM to other models (GMM, LDA, and VAE) and show its superior performance. Finally, we demonstrate our model's interpretability by computing various phonetic properties of ASL through the inspection of our learned model. 

\end{abstract}

\section{Introduction}
American Sign Language (ASL) is the fourth most commonly used language in the United States and is the language most commonly used by Deaf people in the United States and the English-speaking regions of Canada  \cite{ASL-Info-2001}. Unfortunately, until recently, ASL received little research. This is due, in part, to its delayed recognition as a language until William C. Stokoe's publication in 1960 \cite{Stokoe-1960}. Limited data has been a long-standing obstacle to ASL research and computational modeling. The lack of large-scale datasets has prohibited many modern machine-learning techniques, such as Neural Machine Translation, from being applied to ASL. In addition, the modality required to capture sign language (i.e. video) is complex in natural settings (as one must deal with background noise, motion blur, and the curse of dimensionality). Finally, when compared with spoken languages, such as English, there has been limited research conducted into the linguistics of ASL.

We realize a simplified version of Liddell and Johnson's Movement-Hold (MH) Model \cite{Movement-Hold-1989} using a Probabilistic Graphical Model (PGM). We trained our model on ASLing, a dataset collected from three fluent ASL signers. We evaluate our PGM against other models to determine its ability to model ASL. Finally, we interpret various aspects of the PGM and draw conclusions about ASL phonetics. The main contributions of this paper are

\begin{enumerate} [topsep=0pt, itemsep=0pt,partopsep=1ex,parsep=1ex]
    \item A PGM which models the the body configuration in ASL
    \item A video dataset containing over 800 phrases comprising of over 4,000 signs.
\end{enumerate}

\subsection{The Movement-Hold Model}
The Stokoe Model was the first attempt to analyze the internal structure of signs and challenge the notion that signs are holistic iconic gestures. Stokoe analyzed signs using three parameters: location, handshape, and movement. He assumed each parameter was produced simultaneously \cite{txtbk-Stokoe-Model}. 
The foundational Stokoe's model is unable to recognize sequential contrasts occurring between signs because it does not require the parameters of a sign to be produced in a prescribed order \cite{txtbk-Stokoe-Problems}. 

In response to the limitations of the Stokoe Model, Liddell and Johnson published their Movement-Hold (MH) Model. 
The MH Model partitions each sign into sequential hold and movement segments. A bundle of articulatory features is associated with each segment and contains information regarding the segment's handshape, location, palm orientation, and non-manual markers (e.g. facial expressions). 
A segment with no changes to the articulation bundle is a hold segment and can be represented with an H for a full-length hold or an X for a shorter hold. A segment in which some of the articulatory features are transitioning is a movement segment and can be represented with an M. The MH Model recognizes at least nine sign structures, including Hold (H), X M H, and H M X M H, but not H M. The MH Model was novel in its approach to modeling the sequential aspect of signs \cite{txtbk-Movement-Hold-Model}. 

The MH Model is a significant departure from previous models because its basic units of movements and holds are produced sequentially, allowing for precise representation of sequences within signs.

\begin{figure}[H]
\small
    \centering
    \frame{\begin{tabular}{cccccccc}
    $H_1$ & $H_1$ & $H_1$ & $H_2$ & $H_2$ & $E$ & $E$ & $E$\\
    $H_1$ & $H_1$ & $H_1$ & $E$ & $E$ & $E$ & $E$ & $E$\\
    $H_1$ & $H_2$ & $H_3$ & $H_4$ & $H_4$ & $H_4$ & $E$ & $E$
    \end{tabular}}
    \caption{Some example sign structures realized by our model. Each sign is represented in eight frames instead of twenty-five for easier viewing. Each $H_i$ is a distinct body configuration with $E$ being an ending token that is used for padding. The first example illustrates an $HMH$. The second demonstrates an $H$. Finally, the third illustrates an $XMXMXMH$}
    \label{fig:MH_markov}
\end{figure}

\subsection{Prior Works}
While many computational approaches to modeling sign language, including the previously published Interspeech sign language recognition paper \cite{Dreuw-2007} focus on representing signs holistically. A prior approach from speech recognition that gained popularity within sign language modeling is the use of the Hidden Markov Models (HMM) and its variants to model subunits of signs.  Influenced by the MH Model, Theodorakis et al. \cite{Theodorakis-2014} used HMMs to represents signs which they segmented into dynamic and static portions. 
Fang et al. \cite{Fang-2004} modeled Chinese Sign Language using HMMs and temporal clustering. 
Vogler et al. \cite{Vogler-1999} trained HMMs to recognize movement and hold phonemes in an attempt to increase the scalability of ASL recognition systems. 
Bauer et al. \cite{Bauer-2001} take a different approach in their attempt to create an automatic German Sign Language recognition system, using a K-means algorithm to define subunits. 
With all four of these papers focusing on creating a sign language recognition system, there is more focus on recognition success over the phonological model. However, those algorithms do not always identify recognizable phonemes. 
For this reason, building from the ground up, we focus specifically on creating a phonological model and interpreting the model. Additionally, our model is entirely unsupervised, probabilistic, and capable of being trained end-to-end.

\begin{figure*}[ht!]
    \centering
    \subfloat{{\includegraphics[width=5cm]{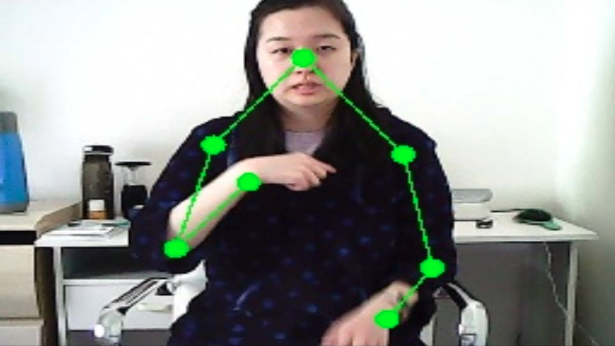} }}%
    \subfloat{{\includegraphics[width=5cm]{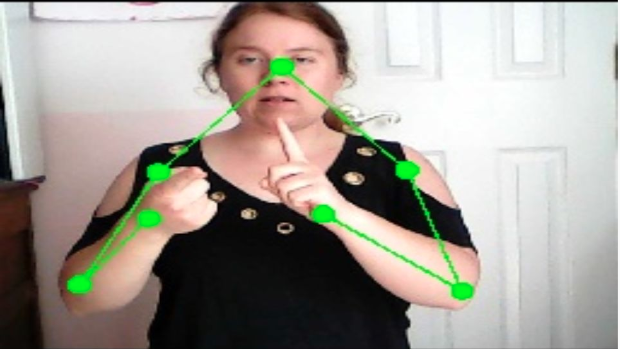} }}%
    \subfloat{{\includegraphics[width=5cm]{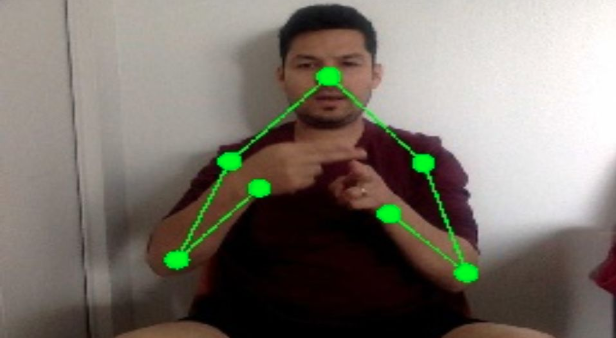} }}%
    \caption{Example frames from our database, ASLing. The green points indicate the features extracted by our Convolutional Pose Machine: the algorithm used to extract body keypoint locations.}%
    \label{fig:example}%
\end{figure*}

\section{Methodology}

We aim to computationally model the phonetic structure within ASL. We restrict ourselves to a simplified form of the MH Model. Our primary deviations from the MH Model are 
\begin{enumerate}[topsep=0pt, itemsep=0pt,partopsep=1ex,parsep=1ex]
    \item we are modelling the global positions (specifically the wrist, elbow, shoulder, and head positions), and
    \item we are explicitly modeling holds and implicitly modelling movement between the holds and compressing it into a single frame.
\end{enumerate}

\subsection{Dataset}
ASLing is an American Sign Language corpus produced by three fluent signers of American Sign Language. Two signers identify as Deaf, and one signer identifies as Hard of Hearing. 
At the time of writing, ASLing contained over 800 phrases. The dataset had over 1,200 signs, with each sign produced an average of about 3 times. The phrases were chosen because they each contained at least one word with a high frequency in English. Each sign was glossed and each phrase was translated into English. Each phrase was tagged with a level of noise (none, low, medium, high, or broken) that described the performance of our pose predictor. The signers were instructed to record with as much of their upper body in frame as possible, in a location with good lighting, and while wearing contrastive clothing. 
The data was collected in phrases instead of individual signs to allow for more natural signing.

\subsection{Feature Extraction and Data Preparation}

The data collected was stored in a video format. Each sign was extracted from the videos (resulting in $M=1515$ signing portions). From each frame of the signing portions, 2D estimates of key points were extracted using a Convolutional Pose Machine \cite{CPM} (specifically the head, shoulders, elbows, and wrist locations). These poses were then normalized (setting the head to the origin and setting the unit to the average distance between the head and shoulders) resulting in $D=14$ features. Finally, the sign were padded to a length of $P=25$ frames with a vector of zeros (as this was the length of the longest sign). Note that the vector of zeros was used as the end token.

\subsection{Phonetic Model}

Through the MH Model, the production of signs can be treated as a sequential process (see Figure \ref{fig:MH_markov} for an example). The entire process can be posed as a Probabilistic Graphical Model (PGM). 
PGMs have several benefits, including their ease of training and interpretability. Additionally, PGMs can be used to generate synthetic data by sampling from the model \cite{pgm}.
We chose specifically to use a Dynamic Bayesian Network to model the sequential aspect of signing. Specifically, let
{\footnotesize
\begin{align*}
    N = &  \text{The number of body configurations prototypes}\\
    M = &  \text{The number of signs}\\
    P = &  \text{The number of frames in each sign}\\
    D = &  \text{The dimensionality of frame features}\\
    \mu_i = &  \text{The $i^{th}$ idealized body configuration prototypes}\\
    \sigma = &  \text{The amount of dispersion around each dimension from the}\\  &\text{idealized body configuration prototypes during sign production}\\
    \pi = &  \text{The probability of each body configuration prototype being at}\\ & \text{the start of a sign}\\
    T_{i, j} = &  \text{The transition probability of moving from body configuration}\\ & \text{prototype $i$ to body configuration prototype $j$}\\
    c^w_i = &  \text{The body configuration prototype chosen for frame $i$ in sign $w$}\\
    x^w_i = &  \text{The observed production of the chosen body configuration}\\ & \text{prototype for frame $i$ in sign $w$}
\end{align*}

}%

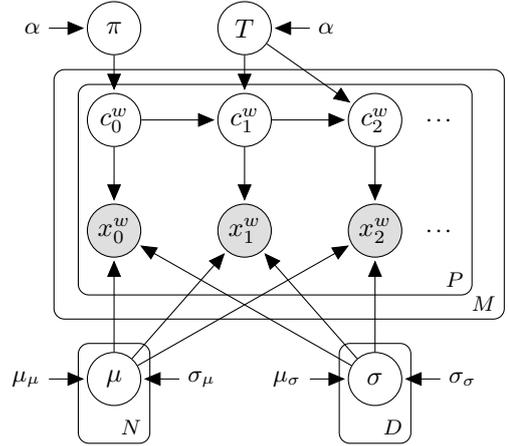
\begin{figure}[H]
    \centering
    \begin{tikzpicture}
        \node[latent] (pi) {$\pi$};
        \node[latent, right=of pi] (T) {$T$};
        \node[left=0.5cm of pi] (alphapi) {$\alpha$};
        \node[right=0.5cm of T] (alphaT) {$\alpha$};
        \node[latent, below=0.5cm of pi] (c0) {$c^w_0$};
        \node[obs, below=.75cm of c0] (x0) {$x^w_0$};
        \node[latent, right=of c0] (c1) {$c^w_1$};
        \node[obs, below=.75cm of c1] (x1) {$x^w_1$};
        \node[latent, right=of c1] (c2) {$c^w_2$};
        \node[obs, below=.75cm of c2] (x2) {$x^w_2$};
        \node[right=0.2 of c2] (ellipsec) {\ldots};
        \node[right=0.2 of x2] (ellipsex) {\ldots};
        \node[above right = 0.2 and 0.2 of ellipsec] (nulla) {};
        \node[below left  = 0.2 and 0.2 of x0] (nullb) {};
        \node[latent, below=1.25cm of x0] (mu) {$\mu$};
        \node[latent, below=1.25cm of x2] (sigma) {$\sigma$};
        \node[left=0.5cm of sigma] (musigma) {$\mu_\sigma$};
        \node[right=0.5cm of sigma] (sigmasigma) {$\sigma_\sigma$};
        \node[left=0.5cm of mu] (mumu) {$\mu_\mu$};
        \node[right=0.5cm of mu] (sigmamu) {$\sigma_\mu$};
        \draw[every loop]
            (musigma) edge node {} (sigma)
            (sigmasigma) edge node {} (sigma)
            (alphapi) edge node {} (pi)
            (mumu) edge node {} (mu)
            (sigmamu) edge node {} (mu)
            (alphaT) edge node {} (T)
            (pi) edge node {} (c0)
            (T) edge node {} (c1)
            (T) edge node {} (c2)
            (c0) edge node {} (c1)
            (c1) edge node {} (c2)
            (c0) edge node {} (x0)
            (c1) edge node {} (x1)
            (c2) edge node {} (x2)
            (mu) edge node {} (x0)
            (mu) edge node {} (x1)
            (mu) edge node {} (x2)
            (sigma) edge node {} (x0)
            (sigma) edge node {} (x1)
            (sigma) edge node {} (x2);
    \plate [] {prototypes} {(mu)} {$N$}; %
    \plate [] {dimensions} {(sigma)} {$D$}; %
    \plate [] {signs} {(c0)(c1)(c2)(x0)(c1)(c2)(ellipsec)(ellipsex)(nulla)(nullb)} {$M$};
    \plate [] {frames} {(c0)(c1)(c2)(x0)(c1)(c2)(ellipsec)(ellipsex)} {$P$};
    \end{tikzpicture}
    \caption{The plate diagram of our PGM. The non-shaded nodes are latent variables. Shaded nodes are observed variables. $\alpha, \mu_\mu, \sigma_\mu, \mu_\sigma, \sigma_\sigma$ are hyperparameters.}
    \label{fig:plate}
\end{figure}

In addition to these, we have the hyperparameters $\alpha, \mu_\mu, \mu_\sigma, \sigma_\mu, \sigma_\sigma$. We specifically chose $\alpha = 1, \mu_\sigma = 1, \sigma_\sigma = 10, \mu_\mu = 0, \sigma_\mu = 10$.
Our full model can then be stated as
\begin{align}
\sigma | \mu_\sigma, \sigma_\sigma  & \sim LogNormal(\mu_\sigma, \sigma_\sigma)\\
T_n  | \alpha & \sim Dir(\alpha \mathbf{1}_N)\\
\mu_0  & = \mathbf{0}_D\\
\mu_{n>0} | \mu_\mu, \sigma_\mu  & \sim \mathcal{N}(\mu_\mu, Diag(\sigma_\mu))\\
\pi | \alpha & \sim Dir(\alpha \mathbf{1}_N)\\
c^w_{f=0}  | \pi & \sim Categorical(\pi)\\
c^w_{f>0} | T, c^w_{f-1} & \sim Categorical(T_{c^w_{f-1}})\\
x^w_f | \mu, c^w_f, \sigma & \sim \mathcal{N}(\mu_{c^w_f}, Diag(\sigma))
\end{align}
This process is shown in the plate diagram in Figure \ref{fig:plate}. Note that the 0th body configuration is defined as a special end phoneme (which is indicated by the 0 feature vector).

\begin{figure}
    \centering
    \subfloat{{\includegraphics[width=3cm, trim={0.7cm 0.7cm 0 0},clip]{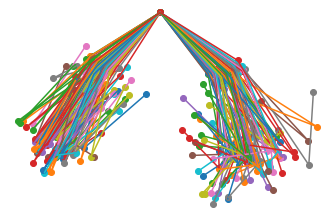} }}%
    \subfloat{{\includegraphics[width=3cm, trim={0.7cm 0.7cm 0 0},clip]{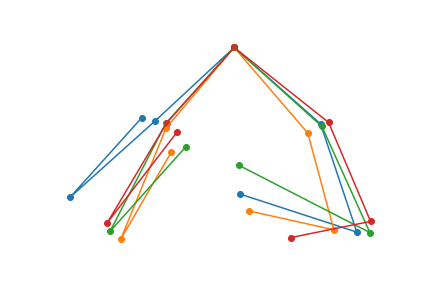} }}%
    \caption{A sampling of body poses exhibited in the data (left) and the learned body configuration prototypes from the 5 body configuration prototype model (right).}
    \label{fig:my_label}
\end{figure}

\subsubsection{Training}

We wish to obtain estimates of $\theta = \pi, T, \mu, \sigma$. With these parameters, one can run the generative process in full and derive maximum a posteriori (MAP) estimates of the model conditioned on data. As we wish only to obtain MAP estimates of these, we will use Expectation Maximization (EM).

To do this, we follow the standard EM procedure:
\begin{enumerate} [topsep=0pt, itemsep=0pt,partopsep=1ex,parsep=1ex]
    \item Calculate the optimal cluster assignments given the data and our current estimation of $\theta$.
    \item Calculate an updated (MAP) estimate of $\theta$ using our cluster assignments and prior on $\theta$. 
\end{enumerate}

To compute the optimal cluster assignments, we calculate $P(Z | X)$:
\begin{align}
    P(c_0^w = i | x_0^w) &\propto e^{\frac{-1}{2} (x_0^w-\mu_i)^T diag(\sigma)^{-1} (x_0^w-\mu_i)} \pi_i\\ 
    P(c_f^w = i | c_{f-1}^w = j, x_f^w) &\propto e^{\frac{-1}{2} (x_f^w-\mu_i)^T diag(\sigma)^{-1} (x_f^w-\mu_i)} T_{j, i}
\end{align}
From this, one can trivially compute the optimal cluster assignments ($Z = c$) given $\theta$ in $\mathcal{O}(MPN)$.

To compute the optimal estimate of $\theta$ given our cluster assignments and priors we use calculate $P(\theta | X, Z)$
\begin{align}
    P(\pi | c_0^w = i) \propto& \frac{\pi_i}{B(\alpha)} \prod_j \pi_j^{\alpha_j - 1}\\ 
    P(T_j | c_f^w = i, c_{f-1}^w = j) \propto& \frac{T_{j, i}}{B(\alpha)} \prod_k \left(T_{j, k}\right)^{\alpha_k - 1} \\
    P(\mu_i | x_f^w, c_f^w = i) \propto& e^{\frac{-1}{2} (x_f^w-\mu_i)^T diag(\sigma)^{-1} (x_f^w-\mu_i)} * \nonumber\\
                                       & e^{\frac{-1}{2} (\mu_i-\mu_\mu)^T diag(\sigma_\sigma)^{-1} (\mu_i^w-\mu_\mu)}\\
    P(\sigma |\mu_i, x, c) \propto& P(x, c, \mu | \sigma) * \nonumber\\
                                  &\prod_i \frac{1}{\sigma_i \sigma_{\sigma_i}} exp\left(-\frac{\ln(\sigma_i) - \mu_{\sigma_i}}{2 \sigma_{\sigma_i}}\right)
\end{align}

where $P(x, c, \mu | \sigma)$ was left unexpanded in the interest of space. These can each be optimized numerically (we chose to optimize these in the order of $\pi, T, \mu, \sigma$). These two steps were repeated until convergence.


\begin{figure}[h]
    \centering
    \includegraphics[height=4cm]{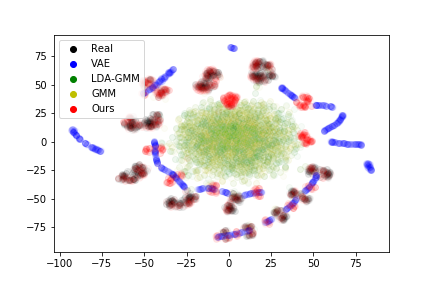}%
    \caption{A T-SNE visualization of the various models presented in this paper}
    \label{fig:tsne}
\end{figure}

\section{Results and Evaluation}

After this model was trained to convergence, we performed the following analysis
\subsection{Evaluation}

\begin{table*}[h]
    \centering
    \begin{tabular}{|c|cccc|}
    \hline
        Model & Ours & GMM & GMM-LDA & VAE \\
    \hline
        Binary Cross Entropy of Descriminator & $\mathbf{0.65 \pm 0.01}$ & $0.05 \pm 0.01$ & $0.03 \pm 0.01$ & $0.04 \pm 0.02$\\
    \hline
    \end{tabular}
    \caption{Binary Cross Entropy (BCE) of a Discriminator trained on distinguishing Synthetic data from Real data. A low BCE means the synthetic generator was unable to adequately fool a discriminator. Note the ability to fool a discriminator corresponds with the ability to produce life-like signs.}
    \label{tab:bce}
	\vspace{-0.6cm}
\end{table*}

We chose to evaluate our model by determining how capable it was of producing signs. We compared our model against several other models:

\begin{enumerate} [topsep=0pt, itemsep=0pt,partopsep=1ex,parsep=1ex]
    \item A GMM-LDA model of sign language
    \item A GMM model of sign language
    \item A (Seq2Seq) Variational Auto Encoder (VAE) \cite{vae}
\end{enumerate}

It should be noted that there was not enough data to successfully train a generative adversarial network\cite{gan} (GAN). The former two models can be seen as ablations of our model.  The VAE was chosen to determine how our model compared with similar deep models. The probabilistic models are described in further detail at the end of this section.


Each of these algorithms were evaluated on its ability to fool a descriminator. We chose to use a simple neural network as a descriminator (specifically a GRU network). This network was trained on the binary problem of distinguishing between generated data (from the generative processes listed above) and real data. The results are shown in Table \ref{tab:bce}.
In this evaluation, our model outperformed all other models by a wide margin. This was likely because the ablated versions were incapable of capturing the dynamic nature of ASL.

In addition to these quantitative experiments, we also used T-SNE\cite{tsne} to project simulated data from each model to two dimensions (shown in Figure \ref{fig:tsne}). From this, we can observe that the LDA-GMM and GMM models underfit (likely due to the high bias of the model). It is also evident that the VAE did not fit the data well. Finally, one can observe that our model produced several signs which did not coincide with real data. This was likely due to our model implicitly modeling movement within the MH Model. Even so, all signs present in the corpus had similar signs generated by our model. This implies that our model was capable of modeling all body configurations present in the corpus.

\subsubsection{GMM-LDA and GMM}
\begin{figure}
	\centering
	\begin{tabular}{cc}
		\frame{\includegraphics[width=0.4\linewidth]{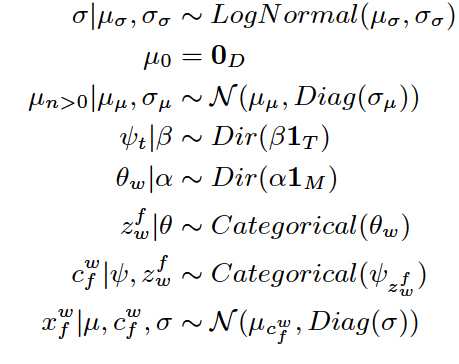}} &
		\frame{\includegraphics[width=0.4\linewidth]{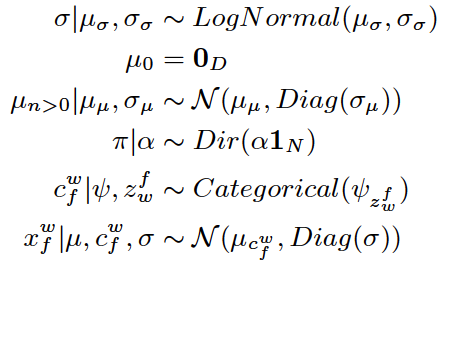}}
	\end{tabular}
	\caption{The model specifications for the LDA-GMM (left) and GMM (right). $f$ is the frame index, $t$ is the topic index, and $T$ is the number of topics. $\alpha, \beta$ are concentration hyperparameters, $\pi$ is the relative frequency of the different body configuration prototypes, and $\psi, \theta$ are the word distributions of the topics and the topic choice for the signs respectively}
	\label{fig:gmmodels}
	\vspace{-0.5cm}
\end{figure}

The GMM-LDA model we used is a topic model\cite{lda} over the body configurations used within a sign. In this model, the topics are distributions of body configuration prototypes and the words are the body configuration prototypes themselves. The GMM model is similar to the GMM-LDA. Their model specifications are given in Figure \ref{fig:gmmodels}.

\subsection{Interpretation}

To interpret the model, we chose to fit a simplified model containing only five body configuration prototypes (including the end state). Note that without loss of generality, this section assumes the signer is right-hand dominant. 
The body configuration prototypes discovered appear to be the following body configurations (where the listed location is the location of articulation for the right hand): end state, head, right chest, left hand, and neutral signing space (i.e. the location used for fingerspelling). We examined $\pi$ to determine that the most common starting position was the right chest area. The least common starting position was the head (we are not counting the end state in the least common starting positions).
In addition, one can determine the average length of each hold. This can be done by examining the self transition probabilities $T[j, j]$ and running Geometric Trials (where the first transition away from the given state is a success):

\begin{align}
\mathbb{E}(\text{length of hold }i) = \frac{1}{1 - T[i, i]}
\end{align}

From this, one can determine most holds last for approximately 8.16 frames (with the standard deviation between the lengths of the different body configuration prototypes being only 0.98 frames). This means that each hold is approximately 800 milliseconds. Note that the holds derived by the non-simplified model are substantially shorter (in the order of 400 milliseconds).
We can also determine the popularity of each body configuration prototype by evaluating the expected number of times each will arise within a sign:
\begin{align}
\mathbb{E}(\text{count of }j) = \sum_{i=0}^{19} P(c_i = j) = \sum_{i=0}^{19} (\pi T^i)[j]
\end{align}

This leads to the conclusion that the ``ending state" will be be in an average of 11 frames. Additionally, the most common pose is the right chest (showing up an average of 2.91 frames per sign) and the least common is the head (showing up an average of 1.86 times per sign).
We can also examine the probability of any given body configuration prototype ending a sign by examining the probability of transitioning to the 0th prototype. Therefore, the right chest is the most common ending body configuration prototype and the left hand is the least common ending body configuration prototype.
Finally, we can examine the amount of variation during production between each joint compared to the body configuration prototype. From examining this, one can observe that the head and shoulder exhibit the least variation. The elbows and left hand exhibit slightly more variation. The right hand exhibits the most variation. This is to be expected as the right hand performs the most motion when signing.

\begin{figure}
    \centering
    \subfloat{{\includegraphics[height=3.5cm]{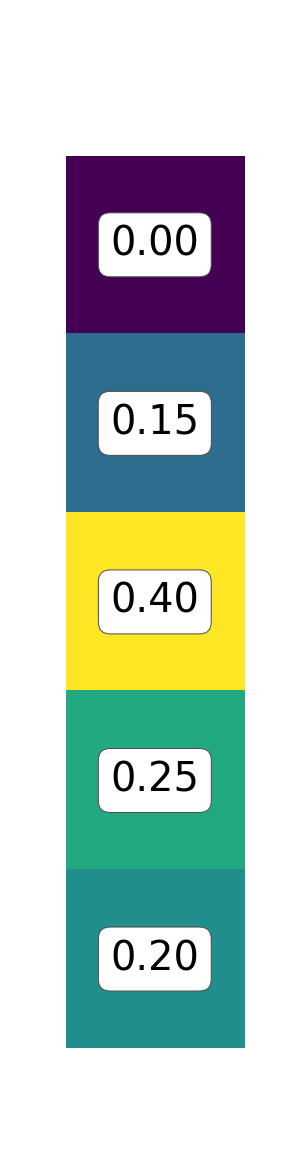} }}%
    \subfloat{{\includegraphics[height=3.5cm]{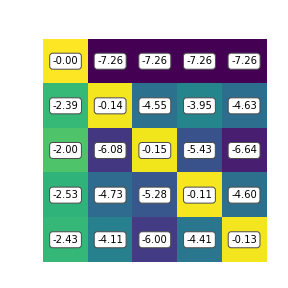} }}%
    \caption{The initial probabilities and (natural) log of transition probabilities learned by our simplified phonetic model. Note the high probability of self transitions. Self transitions mean that one stayed in the body configuration prototype and thus naturally maps to the concept of a Hold (see Figure \ref{fig:MH_markov})}
    \label{fig:probs}
\end{figure}

\section{Conclusions}

In this paper we successfully modeled the phonetic structure of ASL using a simplified Movement-Hold Model. We evaluated our model against multiple ablated versions and a deep counterpart. Finally, we interpreted various aspects of the model to demonstrate its utility and interpretability. Nevertheless, our model did have several limitations. First, we did not explicitly model movement. This likely caused the artifacts seen in Figure \ref{fig:tsne}. Additionally, our model did not model the local movement of the hand (specifically handshape and palm orientation). We plan on not only addressing these issues in future works, but also modeling non-manuals and incorporating 3D to allow for easier interpretation.


\bibliographystyle{IEEEtran}

\bibliography{template}


\end{document}